# Deep Distance Measurement Method for Unsupervised Multivariate Time Series Similarity Retrieval

Susumu Naito
*Corporate Laboratory*
*Toshiba Corporation*
Kawasaki, Japan
susumu.naito.x19@mail.toshiba

Kouta Nakata
*Corporate Laboratory*
*Toshiba Corporation*
Kawasaki, Japan
kouta.nakata.d99@mail.toshiba

Yasunori Taguchi
*Corporate Laboratory*
*Toshiba Corporation*
Kawasaki, Japan
yasunori.taguchi.n49@mail.toshiba

*Abstract*— We propose the Deep Distance Measurement Method (DDMM) to improve retrieval accuracy in unsupervised multivariate time series similarity retrieval. DDMM enables learning of minute differences within states in the entire time series and thereby recognition of minute differences between states, which are of interest to users in industrial plants. To achieve this, DDMM uses a learning algorithm that assigns a weight to each pair of an anchor and a positive sample, arbitrarily sampled from the entire time series, based on the Euclidean distance within the pair and learns the differences within the pairs weighted by the weights. This algorithm allows both learning minute differences within states and sampling pairs from the entire time series. Our empirical studies showed that DDMM significantly outperformed state-of-the-art time series representation learning methods on the Pulp-and-paper mill dataset and demonstrated the effectiveness of DDMM in industrial plants. Furthermore, we showed that accuracy can be further improved by linking DDMM with existing feature extraction methods through experiments with the combined model.

## I. Introduction

Multivariate time series (MTS) data have been accumulated in various domains (industrial plants, medicine, physical activity, etc.). Our particular interest lies in MTS data from industrial plants. In power plants and manufacturing facilities, MTS data from hundreds or thousands of sensors have been collected over time for operations monitoring and maintenance of systems and equipment. There is a growing need for ways to effectively utilize these data for more efficient plant operation. One method is MTS similarity retrieval, which finds MTS data similar to the current state in the historical data. Information about past operating states similar to the current operating states is very useful for analyzing causes after detecting anomalies, evaluating the health of current operating states, and setting new operating conditions.

In the case of industrial plants, although some states are simple (such as uptrends and downtrends), users are most interested in minute differences between the current and past states (such as minute temperature drifts or minute flow fluctuations). Our aim is to improve retrieval accuracy by focusing on such minute differences in the various states of the time series. Here, we define the scope of our study. The objective of our study is to improve retrieval accuracy. Since label information is often unavailable for time series data in industrial plants, we focus on unsupervised settings; supervised settings are outside the scope of this study. Furthermore, because our objective is to improve retrieval accuracy, improving retrieval efficiency is also outside the scope.

A general method for MTS similarity retrieval involves extracting MTS segments (slices of MTS data that last for a short period of time), and then calculating the pairwise distance or metric between the query segment and each segment. In recent years, learning-based methods such as metric learning and representation learning have advanced. For improving MTS retrieval accuracy in unsupervised settings, there are two related approaches. One approach is to use models suitable for capturing MTS features. DUBCNs [1] and UTBCNs [2] employ Long Short-Term Memory Encoder-Decoder (LSTM-ED) and Transformer Encoder-Decoder (Transformer-ED) as their base models, respectively, to improve accuracy. However, since they learn features to reconstruct each segment individually, rather than to capture minute differences between segments, they cannot sufficiently learn the relationships between each segment and the other segments to achieve high accuracy. The other approach to overcome the challenges of learning the relationships is time series representation learning based on contrastive learning, which has made significant progress in recent years [3][4]. The time series representations are obtained as embeddings in Euclidean space and the pairwise distance can be calculated from them in MTS similarity retrieval. The construction of positive and negative samples is essential in contrastive learning. The basic idea for obtaining them is to use MTS segments temporally close to an anchor as positive samples and the other samples as negative samples (Time2State [5], E2USD [6], Triplet-Loss [7], TNC [8], etc.). However, for negative samples, because industrial plants have a wide variety of states, the conventional methods that treat time-distant segments as negative samples cannot learn minute differences between states (Fig. 1). For positive samples, the conventional methods that treat segments temporally close to an anchor as positive samples tend to bias segments ranked high in retrieval toward those temporally close to the query segment. As a result, the period of time covered by the retrieval is narrower, and its performance is degraded for applications such as industrial plants that retrieve MTS over long periods of time. We approach these issues as follows. In order to learn minute differences between states, it is more rational not to use negative samples and to learn features of the differences between anchors and positive samples. It is more rational to obtain positive samples from the entire time series, not just from segments temporally close to the anchor.

Based on the above considerations, we propose the Deep Distance Measurement Method (DDMM) to improve retrieval accuracy. DDMM learns the differences within states by learning features of the differences between anchors and positive samples. DDMM uses two segments, arbitrarily sampled from the entire time series, as a pair of an anchor and a positive sample. In order to use only pairs in which the positive sample is similar to the anchor for the model training, each pair is weighted based on the Euclidean distance within the pair. The weight is set to increase exponentially as the distance decreases. We think that, for retrieval tasks, it is more important to learn the distance space that represents minute differences between anchors and similar positive samples than to keep dissimilar negative samples away from anchors. Therefore, learning for only positive samples based on the distances is effective.

DDMM employs an Autoencoder (AE) [9] as its model. The AE is trained with difference vectors weighted by the weights. The difference vector is the difference within the pair of the anchor and the positive sample. The weight is the above-mentioned weight based on the Euclidean distance within the pair. The reconstruction error of the difference vector is used as the distance measure. Since the difference vectors are trained with the weights, the reconstruction error is smaller for the pair in which the positive sample is more similar to the anchor. We employ the reconstruction error to eliminate query segments with unlearned features and to avoid misrecognition of the pairwise distances (Fig. 2). The combination of an AE and its reconstruction error is often used in real-world anomaly detection, and as an analogy, we employ it in DDMM.

The main contributions of this paper are:

- We propose DDMM to improve retrieval accuracy in unsupervised settings in MTS similarity retrieval. DDMM enables learning minute differences within states in the entire time series and thereby recognition of minute differences between states, which are of interest to users in industrial plants.
- To archive this, DDMM uses a learning algorithm in which each pair of an anchor and a positive sample that are arbitrarily sampled from the entire time series is weighted by the distance within the pair, and the features of the differences within the pairs weighted by the weights are learned. This algorithm allows both learning minute differences within states and sampling pairs from the entire time series.
- We conduct empirical studies on publicly available datasets to analyze the technical effectiveness of DDMM by comparing its performance with those of state-of-the-art methods and assessing the effect of the DDMM architecture.

The rest of this document is organized as follows. Section II discusses related work. Section III describes the details of our method. Sections IV and V describe the experiments and discuss the technical effectiveness of our method.

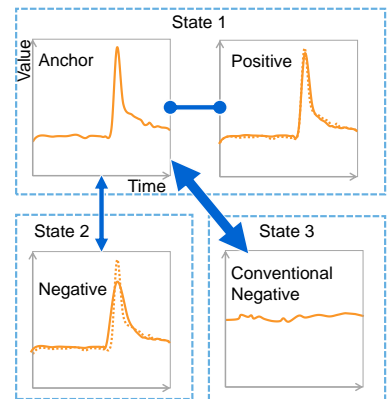

Fig. 1. Conventional negative sample. State 1 and State 2 are slightly different. Conventional negative sample (State 3) cannot distinguish between State 1 and State 2.

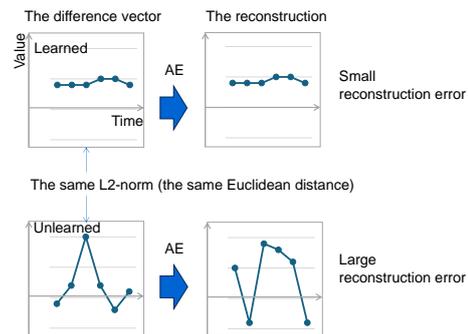

Fig. 2. Role of the reconstruction error. The L2 norm of the difference vector expresses the Euclidean distance within a pair. Even if the L2 norm of the difference vector with unlearned features is the same as that with learned features, the reconstruction error is large and the similarity is rejected.

## II. RELATED WORK

Here we discuss two approaches related to improving MTS retrieval accuracy in unsupervised settings.

One approach is to use models suitable for capturing MTS features. The models have the encoder-decoder structure for capturing the MTS features. The encoder output and the decoder input are connected to the metric space. DUBCNs [1] and UTBCNs [2] employ LSTM-ED and Transformer-ED as their respective base models to improve accuracy. However, because they learn features to reconstruct each segment individually, rather than to capture minute differences between segments, the relationships between each segment and the other segments cannot be sufficiently learned to achieve high accuracy.

The other approach to overcome the challenges of learning the relationships is time series representation learning based on contrastive learning. Time series representation learning assumes classification and does not explicitly address similarity retrieval. However, we think that classification and similarity retrieval are essentially the same problem, although similarity retrieval requires that among segments in the same state as the query segment, the segment more similar to the query segment should be ranked higher. Dozens of methods have been proposed [3][4]. The construction of positive and negative samples is essential in contrastive learning. The basic idea to obtain them is to use MTS segments temporally close to an anchor as positive samples and the other samples as negative samples. Here are the state-of-the-art examples: Time2State [5] randomly samples multiple sets of consecutive segments from the time series, with each segment as an anchor, its consecutive segments as positive samples, and other randomly sampled segments as negative samples. E2USD [6], an evolution of Time2State, further selects genuinely dissimilar negative pairs by evaluating seasonal and trend similarities for each pair of randomly sampled segments and retaining the least similar ones as negative samples. Triplet-Loss [7] treats segments in a subseries sampled from the timeseries as an anchor and positive samples, and those in another subseries as negative samples. TNC [8] treats temporal neighborhoods as positive samples through the Augmented Dickey-Fuller test and treats the non-neighborhoods as negative samples with smaller weights based on Positive-Unlabeled learning. In addition, there are methods that use two augmentations from one segment as positive pairs and those from the other segments as negative samples. TS-TCC [10] applies jitter-and-scale and permutation-and-jitter augmentations to have a positive pair for each segment. TS2Vec [11] applies timestamp masking and random cropping augmentations, and treats representations at the same timestamp as a positive pair.

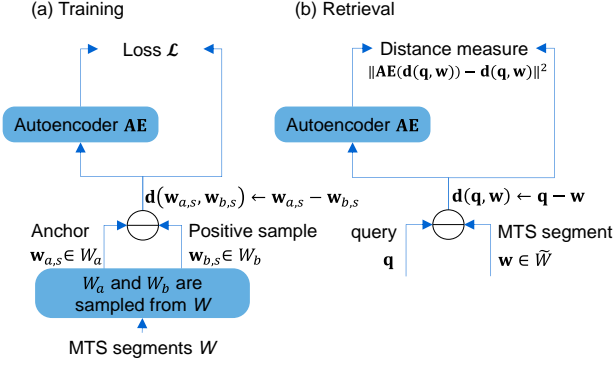

Fig. 3. DDMM configuration. (a) Training: A set of $S$ anchors ($W_a$) and a set of $S$ positive samples ($W_b$) are sampled from the training MTS segments $W$. A pair of $\mathbf{w}_{a,s}$ and $\mathbf{w}_{b,s}$ expresses a pair of the $s$-th anchor and the corresponding positive sample. The **AE** is trained with the difference vectors $\{\mathbf{d}(\mathbf{w}_{a,s}, \mathbf{w}_{b,s})\}_{s=1}^{S}$ weighted by their Euclidean distances. (b) Retrieval: The difference vector $\mathbf{d}(\mathbf{q}, \mathbf{w})$ between a given query MTS segment $\mathbf{q}$ and a segment $\mathbf{w}$ of the test MTS segments $\widetilde{W}$ is input to the **AE** to obtain the reconstruction error as the distance measure.

However, for negative samples, because industrial plants have a wide variety of states, the conventional methods that treat time-distant segments as negative samples cannot learn minute differences between states (Fig.1). In addition, if segments temporally close to an anchor are treated as negative samples to learn the minute difference, positive samples can be misidentified as negative samples. For positive samples, the conventional methods that treat segments temporally close to an anchor as positive samples tend to bias segments ranked high in retrieval toward those temporally close to a query segment. The period of time covered by the retrieval is narrower, and its performance is degraded for applications such as industrial plants that retrieve MTS over long periods of time. In addition, the augmentation methods are difficult to apply in industrial plants. Because sensor values vary in complex ways, appropriate augmentation is not intuitive.

To address these challenges, in order to learn minute differences within states and thereby to recognize minute differences between states, our method does not use negative samples and learns features of the differences between anchors and positive samples. Positive samples are sampled from the entire time series, not just segments temporally close to the anchor.

## III. METHOD

### A. Problem Formulation

We formulate an MTS similarity retrieval problem. We denote multivariate time series as a sequence of data points: $\{\mathbf{x}_1, ..., \mathbf{x}_T\}$, where $\mathbf{x}_t \in \mathbb{R}^m$ is a column vector of values of $m$ variables at time $t$. MTS contains $m$ variables at each time instant. We define $\mathbf{w}_t$ as an MTS segment of length $K$ at time $t$.

$$\mathbf{w}_t = \begin{bmatrix} \mathbf{x}_{t-(K-1)}^\top & \cdots & \mathbf{x}_{t-1}^\top & \mathbf{x}_t^\top \end{bmatrix}^\top, \quad (1)$$

where $\cdot^\top$ is the transpose of a matrix. An MTS segment $\mathbf{w}_t$ is an $mK$-dimensional column vector. We define $W$ as a set of $\mathbf{w}_t$.

$$W = \{\mathbf{w}_K, \mathbf{w}_{K+1}, ..., \mathbf{w}_T\}. \quad (2)$$

The MTS similarity retrieval problem is to find the top $k$ MTS segments from $W$ that are most similar to a given query MTS segment $\mathbf{q}$ based on the pairwise distance $\Delta(\mathbf{q}, \mathbf{w}_t)$ expressed by the distance measure. If $\Delta(\mathbf{q}, \mathbf{w}_{t1}) < \Delta(\mathbf{q}, \mathbf{w}_{t2})$, then $\mathbf{w}_{t1}$ is ranked higher than $\mathbf{w}_{t2}$.

### B. Deep Distance Measurement Method

Fig. 3(a) shows the configuration of DDMM in training. Algorithm 1 shows the training algorithm. In the training, a set of $S$ anchors ($W_a$) and a set of $S$ positive samples ($W_b$) are sampled from the training MTS segments $W$, where $S$ is the batch size, $W_a = \{\mathbf{w}_{a,s}\}_{s=1}^{S}$, $W_b = \{\mathbf{w}_{b,s}\}_{s=1}^{S}$, and $\mathbf{w}_{a,s}$ and $\mathbf{w}_{b,s}$ are the $s(=1,...,S)$-th element of $W_a$ and $W_b$, respectively. A pair of $\mathbf{w}_{a,s}$ and $\mathbf{w}_{b,s}$ expresses a pair of the $s$-th anchor and the corresponding positive sample.

In this paper, we use randomly sampled segments from $W$ as $W_a$ and $W_b$ in order to efficiently train the **AE**. This random selection strategy is effective for MTS data that contain many segments in the same state as any given anchor.

Next, the difference vector $\mathbf{d}(\mathbf{w}_{a,s}, \mathbf{w}_{b,s})$ between $\mathbf{w}_{a,s}$ and $\mathbf{w}_{b,s}$ is calculated.

$$\mathbf{d}(\mathbf{w}_{a,s}, \mathbf{w}_{b,s}) = \mathbf{w}_{a,s} - \mathbf{w}_{b,s}. \quad (3)$$

$\mathbf{d}(\mathbf{w}_{a,s}, \mathbf{w}_{b,s})$ is input to the autoencoder **AE**. The **AE** is trained to minimize the loss $\mathcal{L}$ as follows:

$$\mathcal{L} = \frac{1}{\sum_{s=1}^{S} w(\|\mathbf{d}(\mathbf{w}_{a,s}, \mathbf{w}_{b,s})\|)} \cdot \sum_{s=1}^{S} w(\|\mathbf{d}(\mathbf{w}_{a,s}, \mathbf{w}_{b,s})\|) \\ \cdot \left\| \mathbf{AE}\left(\mathbf{d}(\mathbf{w}_{a,s}, \mathbf{w}_{b,s})\right) - \mathbf{d}(\mathbf{w}_{a,s}, \mathbf{w}_{b,s}) \right\|^2, \quad (4)$$

where $w(d) = \frac{1}{d^2 + \delta}$ is the weight function, $\delta$ is an arbitrarily small constant for zero-division prevention, and $\|\cdot\|$ is L2-norm. The weight $w(\|\mathbf{d}(\mathbf{w}_{a,s}, \mathbf{w}_{b,s})\|)$ exponentially increases as the Euclidean distance within the pair, i.e., $\|\mathbf{d}(\mathbf{w}_{a,s}, \mathbf{w}_{b,s})\|$ decreases. In (4), the inverse of the sum of all weights in the batch is multiplied as the normalization constant to reduce the extreme fluctuation in losses from batch to batch.

Fig. 3(b) shows the configuration of DDMM in retrieval. Algorithm 2 shows the retrieval algorithm. In the retrieval, the difference vector $\mathbf{d}(\mathbf{q}, \mathbf{w})$ between a given query MTS segment $\mathbf{q}$ and a MTS segment $\mathbf{w}$ of the test MTS segments $\widetilde{W}$ is input to the **AE** to obtain the reconstruction error DDM$(\mathbf{q}, \mathbf{w})$ as the distance measure.

$$\text{DDM}(\mathbf{q}, \mathbf{w}) = \left\| \mathbf{AE}(\mathbf{d}(\mathbf{q}, \mathbf{w})) - \mathbf{d}(\mathbf{q}, \mathbf{w}) \right\|^2. \quad (5)$$

For all elements of $\widetilde{W}$, the reconstruction errors are calculated to obtain a set of DDMs $D(= \{\text{DDM}(\mathbf{q}, \mathbf{w}) | \mathbf{w} \in \widetilde{W}\})$. The smaller DDM$(\mathbf{q}, \mathbf{w})$ is, the more similar $\mathbf{w}$ is to $\mathbf{q}$. Sorting all elements of $D$ in ascending order yields top $k$ ranking.

---

**Algorithm 1 training algorithm**

**Input:** Training MTS segments $W = \{\mathbf{w}_1, \dots, \mathbf{w}_T\}$,
Number of epochs $E$, Number of iterations $I$, Batch size $S$
**Output:** Trained **AE**
**AE** ← Initialize weights
**for** $e = 1$ to $E$ **do**  # epoch
  **for** $i = 1$ to $I$ **do**  # iteration
    $W_a \leftarrow S$ anchors from $W$  # $W_a = \{\mathbf{w}_{a,s}\}_{s=1}^S$
    $W_b \leftarrow S$ positive samples from $W$  # $W_b = \{\mathbf{w}_{b,s}\}_{s=1}^S$

    define $\mathbf{d}(\mathbf{w}_{a,s}, \mathbf{w}_{b,s})$ as $\mathbf{d}(\mathbf{w}_{a,s}, \mathbf{w}_{b,s}) = \mathbf{w}_{a,s} - \mathbf{w}_{b,s}$
    define $w(d)$ as $w(d) = \frac{1}{d^2 + \delta}$

$$\mathcal{L} \leftarrow \frac{1}{\sum_{s=1}^S w(\|\mathbf{d}(\mathbf{w}_{a,s}, \mathbf{w}_{b,s})\|)} \cdot \sum_{s=1}^S w(\|\mathbf{d}(\mathbf{w}_{a,s}, \mathbf{w}_{b,s})\|) \cdot \|\mathbf{AE}(\mathbf{d}(\mathbf{w}_{a,s}, \mathbf{w}_{b,s})) - \mathbf{d}(\mathbf{w}_{a,s}, \mathbf{w}_{b,s})\|^2$$

    **AE** ← Update weights to minimize $\mathcal{L}$
  **end for**
**end for**

---

**Algorithm 2 retrieval algorithm**

**Input:** Test MTS segments $\widetilde{W}$, Query MTS segment $\mathbf{q}$
**Output:** Top $k$ ranking
$D \leftarrow \{\}$  # Initialize set of DDMs $D$ as an empty set
**for** $\mathbf{w}$ in $\widetilde{W}$ **do**
  DDM$(\mathbf{q}, \mathbf{w}) \leftarrow \|\mathbf{AE}(\mathbf{d}(\mathbf{q}, \mathbf{w})) - \mathbf{d}(\mathbf{q}, \mathbf{w})\|^2$
  $D \leftarrow D \cup \{\text{DDM}(\mathbf{q}, \mathbf{w})\}$
**end for**
Sort all elements of $D$ in ascending order

---

## IV. EXPERIMENTAL SETUP

This section describes the datasets, pre-processing, training and test data, hyperparameters, implementation, query, retrieval, and performance metrics used in our experiments.

### A. Public Datasets

We used three publicly available datasets for our empirical study. Table I summarizes the characteristics of the datasets.

*Pulp-and-paper mill dataset.* This dataset [1] [12] comprises continuous measurement data collected from sensors installed in paper manufacturing machines in a pulp-and-paper mill. Several sensors are placed in different parts of the machine. These sensors measure both raw materials (e.g., amount of pulp fiber, chemicals) and process variables (e.g., blade type, couch vacuum, rotor speed). The sampling period is 2 minutes and the measurement time is 29 days. Paper manufacturing is a continuous rolling process, during which paper sometimes breaks. The data include 124 breaks. The label is '1' at the time of the break and '0' otherwise. A '1' appears at the break time, after which data are missing for up to approximately 1 hour. A '1' appears only once for each break.

*EEG eye state dataset.* This dataset [2] consists of continuous EEG measurement data collected with the Emotiv EEG Neuroheadset. The sampling period is 7.8 milliseconds and the measurement time is 117 seconds. The eye state was detected via a camera during the EEG measurement and manually added to the data file after the video frames were analyzed. The label is '1' for the eye-closed state and '0' for the eye-open state.

*PAMAP2 dataset.* The PAMAP2 physical activity monitoring dataset [3] contains data on 18 different physical activities such as 'lying' 'walking', 'running', 'ascending stairs', and 'rope jumping' performed by 9 subjects wearing 3 inertial measurement units and a heart rate monitor. The sampling period is 10 milliseconds (100 Hz). The label indicates the ID number corresponding to each physical activity. We used the data (subject101.dat) from the first of the 9 subjects.

TABLE I. Benchmarked Datasets.

| Dataset | # of sensors | # of instances | # of classes |
|---|---|---|---|
| Pulp-and-paper mill | 61 | 18398 | 2 |
| EEG eye state | 14 | 14980 | 2 |
| PAMAP2 | 52 | 376417 | 18 |

### B. Pre-processing

*Data cleansing.* The sensor values in the EEG eye state dataset include large pulse signals considered to be measurement noise, which interfere with the performance evaluation. Therefore, we removed these pulse signals by eliminating rows in the data where any of the sensor values contain pulse signals (see the Appendix). In the PAMAP2 dataset, the sensor 'heart rate' was excluded because almost all its values were NaN. For the remaining data, rows with NaN for any of the sensor values were removed. As a result, the number of sensors was reduced from 52 to 51, and the number of instances from 376,417 to 373,161.

*Down-sampling.* To reduce computational costs, the PAMAP2 dataset was down-sampled from the original 100 Hz to 10 Hz. We think that the original 100 Hz sampling frequency is too high to capture physical activity, with 10 Hz being sufficient, and the original waveform information is not lost even if down-sampled. For down-sampling, we used resample().min() from the pandas library on python.

*Normalization.* The data are normalized by scaling the range (minimum value, maximum value) of each sensor value to the range (0, 1).

---

[1] https://docs.google.com/forms/d/e/1FAIpQLSdyUk3lfDl7I5KYK_pw285LCApc-_RcoC0Tf9cnDnZ_TWzPAw/viewform
[2] https://archive.ics.uci.edu/ml/datasets/EEG+Eye+State
[3] http://archive.ics.uci.edu/ml/datasets/pamap2+physical+activity+monitoring

## C. Training and Test Data

The training and test data were identical. There are both online real-time and offline applications for MTS similarity retrieval, and both are in demand. Our experimental conditions correspond to offline applications. Since online real-time applications where the training and test data are different tend to complicate performance evaluation, we selected the offline applications in this paper. We set the segment-length $K$ to 10 and the stride to 1 for all datasets.

## D. DDMM Hyperparameters and Implementation

We set the hyperparameters of the AE to empirical values. The AE was a fully connected AE with three intermediate layers. The configuration was $1 \rightarrow 0.75 \rightarrow 0.5 \rightarrow 0.75 \rightarrow 1$, where the numbers indicate the ratio of the number of nodes in each layer to the number of dimensions in the input layer. The activation functions in the intermediate and output layers were sigmoid functions. The training data were randomly divided into training : validation = 4 : 1. The training epochs were 5000 for the Pulp-and-paper mill dataset, 50 for the EEG eye state dataset, and 100 for the PAMAP2 dataset in all experiments. Because paper breaks are rare events, we used a large number of epochs for the Pulp-and-paper mill dataset to increase the probability of including paper breaks in randomly sampled segments (see Sec. III. D). The packages and versions used for the algorithm implementation were: python 3.9.18, tensorflow 2.11.1, keras 2.11.0, numpy 1.26.0, pandas 2.1.1, scikit-learn 1.3.1, scipy 1.11.3, and tslearn 0.6.3. As an optimizer, we used Adam operator with a learning rate of 0.001. All experiments were conducted on an NVIDIA GeForce GTX 1080 Ti 11GB GDDR5X GPU. Using the above hyperparameters, the training times of our model were 729 seconds for the Pulp-and-paper mill dataset, 3.44 seconds for the EEG eye state dataset, and 19.7 seconds for the PAMAP2 dataset.

## E. Query

For the Pulp-and-paper mill dataset, we used all segments of all 124 breaks as queries (i.e., 124 queries in total). Each query was $\mathbf{w}_t$ in (1) when $t$ was the time of the break (label '1'). For the EEG eye state dataset, we used 500 randomly sampled segments as queries. For the PAMAP2 dataset, we used 500 randomly sampled segments as queries, except during the transition to the next movement (activityID = 0), which does not refer to a specific physical activity.

## F. Retrieval

To prevent extracting a segment identical to a query segment from the test data in the retrieval, segments in the test data that overlapped even partially with the query segment were excluded from the test data, and then the test data were retrieved. We obtained our top $k$ rankings by sorting the pairwise distances (the reconstruction errors $D$) in ascending order.

## G. Performance Metrics

We used three evaluation metrics, namely, mean average precision at top $k$ (MAP@k) [4], precision at top $k$ (Precision@k) [5], and recall at top $k$ (Recall@k). The label at time $t$ in the dataset was set to represent the state of segment $\mathbf{w}_t$. If the label of a segment extracted in retrieval is the same as the label of a query segment, then it is correct; otherwise it is incorrect. Since users in industrial plants focus on segments ranked high in retrieval, achieving high accuracy at the top of the rankings is important. Therefore, we employed $k = 1$ and $k = 10$ for MAP@k.

## V. EXPERIMENTS AND RESULTS

We evaluated performance and technical effectiveness of DDMM. Note that MAP@k, Precision@k, and Recall@k values shown in the figures and tables are averages over three experiments using different random number seeds.

### A. Overall Performance

To show the overall performance of DDMM, we compared it with baseline methods: k-nearest neighbors by Euclidean distance (EU) as the simplest method; AE; LSTM-ED [13] as the base model of DUBCNs; UTBCNs, a state-of-the-art method of unsupervised MTS similarity retrieval; and state-of-the-art time series representation learning methods (TNC, TS2Vec, Time2State, and E2USD). AE has been successfully used in recent work on anomaly detection for MTS in scaled-down versions of industrial plants [14]. For AE and LSTM-ED, we used the encoder output of AE and the output of the last LSTM unit of LSTM-Encoder as MTS representations (embeddings in Euclidean space), respectively. For AE, LSTM-ED, and the state-of-the-art time series representation learning methods, we used the Euclidean distance between embeddings as the pairwise distance. We obtained top $k$ rankings by sorting the pairwise distances in ascending order.

We also evaluated a combined model (AE + DDMM) that uses representations from the baseline method AE as input to DDMM. We think that learning minute differences between representations, rather than those between segments, can be more accurate in capturing minute differences within states. In AE + DDMM, MTS segments are converted to embeddings by the trained AE, and DDMM is then applied to these embeddings, not to the MTS segments.

Following [14], the baseline method AE configuration was $1 \rightarrow 0.5 \rightarrow 1$. The number of dimensions in the input layer was that of MTS segment ($= m \times K$). The activation functions in the intermediate and output layers were sigmoid functions. We used the output of the intermediate layer of the AE as the encoder output. The LSTM-ED configuration was time step $= K$ and the latent dimensions $= 0.5 \times m \times K$. For UTBCNs, TNC, TS2Vec, Time2State, and E2USD, we used publicly available codes. UTBCNs use Hamming distance as the metric and output representations as binary codes. Since the larger the binary length, the higher the accuracy [2], we set it to 1204 bits, which is the memory limit of the GPU we used. Because of overfitting problem with UTBCNs as described in [2], we employed the output at epochs (10 to 20) with the highest accuracy for each dataset in epochs 0 to 100. For TNC, TS2Vec, Time2State, and E2USD, we employed the output at 20 epochs for all datasets. Each method has almost the same accuracy in epochs 20 to 100 and no overfitting was observed. For TS2Vec, which treats MTS in instances and not in segments, we created segments from embeddings in instances and then calculated the pairwise distances.

Tables II and III show the experimental results of MAP@1 and MAP@10. DDMM and AE + DDMM showed high accuracy for the three datasets. In particular, they significantly outperformed the baseline methods for the Pulp-and-paper mill dataset. We think that this is because DDMM captured minute differences within states in the entire time series. AE + DDMM showed better accuracy than DDMM. We think that this is because learning minute differences between representations, rather than those between segments, allowed DDMM to more accurately capture minute differences within states.

---

[4] https://www.evidentlyai.com/ranking-metrics/mean-average-precision-map
[5] https://www.evidentlyai.com/ranking-metrics/precision-recall-at-k#precision-at-k

For the EEG Eye State dataset, UTBCNs and E2USD showed lower accuracy than the other methods. UTBCNs were less accurate than LSTM-ED. Compared to LSTM-ED, Transformer-ED, the base model of UTBCNs, captures not only temporal correlations but also inter-sensor correlations [2]. We think that UTBCNs failed to sufficiently capture inter-sensor correlations for this dataset. E2USD was less accurate than Time2State. E2USD is an evolution of Time2State and uses the selection strategy of negative pairs based on seasonal and trend similarities. We think that the strategy did not work effectively for EEG signals with short measurement times.

For the PAMAP2 dataset, all methods achieved high accuracy. We found that physical activity signals in this dataset had large differences between states.

The rest of the discussion in this paper focuses on the Pulp-and-paper mill dataset, as it clearly showed performance differences among the methods and is particularly relevant to our interest in MTS data from industrial plants.

For the Pulp-and-paper mill dataset, EU and AE showed the same accuracy. In contrast, AE + DDMM was more accurate than DDMM. We think that although AE captured features of segments, its representations were not those that emphasized differences between states in Euclidean space. UTBCNs were more accurate than LSTM-ED. We think that this is because UTBCNs captured inter-sensor correlations, which are important in industrial plants where many sensors are closely linked and fluctuate simultaneously. The state-of-the-art time series representation learning methods (TNC, TS2Vec, Time2State, and E2USD) were more accurate than UTBCNs. We think that this is because they actively learned the differences between states through contrastive learning. As described above, DDMM and AE + DDMM showed clear superiority over the baseline methods. In addition, TS2Vec, an augmentation method, showed accuracy comparable to the other time series representation learning methods. We think that the augmentations of TS2Vec were effective for this dataset.

Fig. 4 shows the average values of Precision@k and Recall@k over all 124 queries for the Pulp-and-paper mill dataset. For Precision@k and Recall@k, DDMM and AE + DDMM outperformed the other methods at $k < 50$. The curves of Precision@k for DDMM and AE + DDMM tended to decrease as $k$ increased, whereas those for the other methods remained roughly constant. Clearly, the trends for DDMM and AE + DDMM were different from those for the other methods. We think that this is because DDMM focuses on minute differences within states, since the smaller $k$ is, the more important minute differences between a query segment and the other segments are. Since users in industrial plants pay attention to segments ranked high in retrieval, it is important that high accuracy is obtained at the top of the rankings.

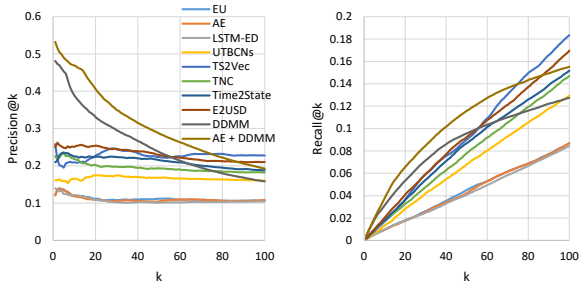

Fig. 4. Precision@k and Recall@k for Pulp-and-paper mill dataset.

TABLE II. Experimental results of MAP@1.

| Methods | Pulp-and-paper mill | EEG eye state | PAMAP2 |
|---|---|---|---|
| EU | 0.121 | 0.922 | 0.986 |
| AE | 0.121 | **0.939** | 0.986 |
| LSTM-ED | 0.140 | 0.908 | **0.987** |
| UTBCNs | 0.167 | 0.601 | 0.920 |
| TNC | 0.234 | 0.823 | 0.977 |
| TS2Vec | 0.237 | 0.878 | 0.963 |
| Time2State | 0.234 | 0.817 | 0.985 |
| E2USD | 0.250 | 0.725 | 0.949 |
| DDMM | 0.470 | 0.921 | 0.965 |
| AE + DDMM | **0.522** | **0.939** | 0.963 |

TABLE III. Experimental results of MAP@10.

| Methods | Pulp-and-paper mill | EEG eye state | PAMAP2 |
|---|---|---|---|
| EU | 0.171 | 0.926 | 0.987 |
| AE | 0.171 | 0.936 | 0.987 |
| LSTM-ED | 0.173 | 0.904 | **0.988** |
| UTBCNs | 0.266 | 0.678 | 0.918 |
| TNC | 0.281 | 0.845 | 0.981 |
| TS2Vec | 0.299 | 0.911 | 0.968 |
| Time2State | 0.277 | 0.838 | 0.986 |
| E2USD | 0.305 | 0.773 | 0.952 |
| DDMM | 0.517 | 0.926 | 0.961 |
| AE + DDMM | **0.562** | **0.937** | 0.962 |

*B. The Effect of Learning Minute Differences within States*

We investigated the effect of learning minute differences within states. Fig. 5 shows a comparison model with DDMM, which learns not features of minute differences within pairs, but features of anchors and positive samples. The comparison model is a simple model of distance metric learning. $\mathbf{AE}^{(1)}$ is an encoder and its configuration is $1 \to 0.75 \to 0.5 \to 0.75 \to 1$, i.e., the same configuration as the AE we used in DDMM. We employed this configuration to eliminate the differences between the configurations of the comparison model and the DDMM model and to focus the evaluation on the presence or absence of learning of minute differences. The loss $\mathcal{L}_1$ and the distance measure $\mathrm{DC}_1(\mathbf{q}, \mathbf{w})$ were as follows:

$$\mathcal{L}_1 = \frac{1}{\sum_{s=1}^{S} w(\|\mathbf{d}(\mathbf{w}_{a,s}, \mathbf{w}_{b,s})\|)} \cdot \sum_{s=1}^{S} w(\|\mathbf{d}(\mathbf{w}_{a,s}, \mathbf{w}_{b,s})\|)$$
$$\cdot \|\mathbf{AE}^{(1)}(\mathbf{w}_{a,s}) - \mathbf{AE}^{(1)}(\mathbf{w}_{b,s})\|^2, \quad (6)$$

$$\mathrm{DC}_1(\mathbf{q}, \mathbf{w}) = \|\mathbf{AE}^{(1)}(\mathbf{q}) - \mathbf{AE}^{(1)}(\mathbf{w})\|^2. \quad (7)$$

Compared to (4) and (5), in (6), the encoder $\mathbf{AE}^{(1)}$ is trained to minimize the distances within the encoded pairs. In (7), the distance within the encoded pair is used as the distance measure. The experimental conditions for the comparison model were the same as those for the DDMM. We used the Pulp-and-paper mill dataset.

Table IV shows MAP@1 and MAP@10. Compared to DDMM, the comparison model showed a significant decrease in accuracy. This notable decline indicates that the main reason for DDMM's high accuracy is the effect of learning minute differences within states and that it is essential to learn them in order to improve retrieval accuracy.

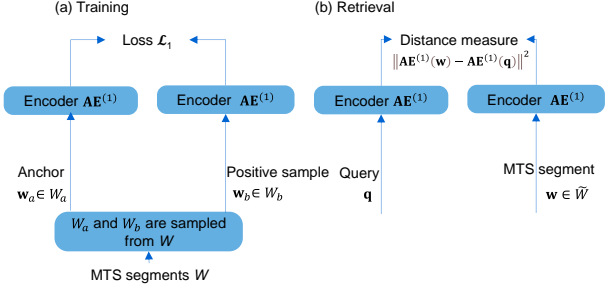

Fig. 5. The comparison model with DDMM. This model learns not features of minute differences within pairs, but features of anchors and positive samples.

TABLE IV. Experimental results of DDMM and the comparison model for Pulp-and-paper mill dataset.

| Methods | MAP@1 | MAP@10 |
|---|---|---|
| DDMM | **0.470** | **0.517** |
| Comparison model | 0.167 | 0.214 |

### C. The Effect of Sampling Positive Samples from the Entire Time Series

The conventional time series representation learning methods treat segments temporally close to an anchor as positive samples. In contrast, DDMM uses segments sampled from the entire time series as positive samples. We investigated this effect. Fig. 6 shows histograms of the time differences between all query segments and the corresponding top-10 segments in retrieval for the Pulp-and-paper mill dataset. Fig. 7 shows those between them and the correct segments in the top-10 segments. The time difference is $t1 - t2$ for a query segment $\mathbf{w}_{t1}$ and a top-10 segment $\mathbf{w}_{t2}$. As shown in Table I, the number of instances in this dataset is 18398, and thus the range of the time difference is -18398 to 18398. As shown in Fig. 6, the time differences for the time series representation learning methods were concentrated near 0. The top-10 segments were biased toward segments temporally close to the query segment. In contrast, DDMM and AE + DDMM had widely distributed time differences, and as shown in Fig. 7, they widely extracted the correct segments from the entire time series. This indicates the effect of sampling positive samples from the entire time series. This characteristic is advantageous for applications, such as those in industrial plants that retrieve MTS over long periods of time.

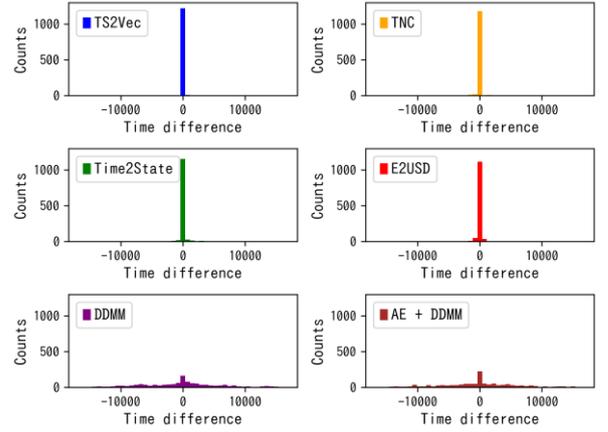

Fig. 6. Histograms of the time differences between all query segments and the corresponding top-10 segments in retrieval for Pulp-and-paper mill dataset.

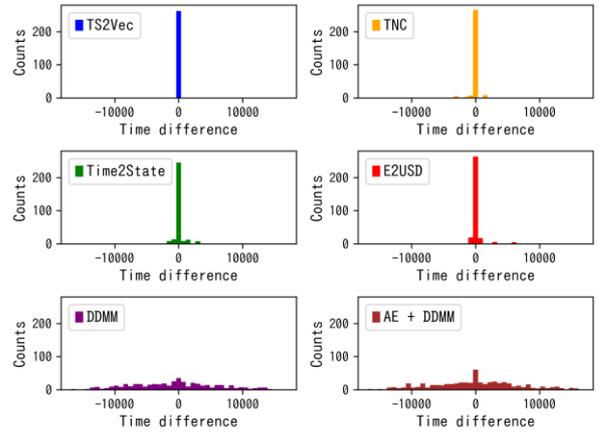

Fig. 7. Histograms of the time differences between all query segments and the correct segments in the top-10 segments for Pulp-and-paper mill dataset.

### D. The Effect of the Random Selection Strategy

In this paper, we used the random selection strategy for sampling positive samples in order to efficiently train the AE in DDMM. This strategy is effective for MTS data that contain many segments in the same state as any given anchor. We investigated the effectiveness of this strategy for the Pulp-and-paper mill dataset, where the paper break, one of the two classes, is a rare event. Fig. 8 shows MAP@10 for varying the ratio of the number of randomly sampled pairs to the number of combinations of two segments in the entire time series. The ratio was varied by varying the number of epochs. In 1000 epochs, 5.4% of the combinations are trained. MAP@10 increased as the number of epochs increased. On the other hand, even 5.4% of the combinations (1000 epochs) had retrieval accuracy close to that at 100% of the combinations (18362 epochs). Although the random selection strategy is data-dependent, it was effective to train the AE efficiently for this dataset. In industrial plants, operations are essentially continuations of the same operating states or repetitions of the same operating patterns. Even anomalous states often persist for some period of time. Therefore, MTS data tend to contain many segments in the same state. Random selection can be an effective strategy.

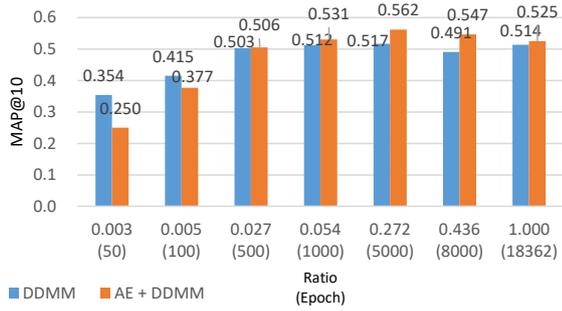

Fig. 8. MAP@10 vs. epochs for Pulp-and-paper mill dataset.

*E. Segment-length K Dependence*

The selection of the segment-length $K$ is the most difficult problem in unsupervised settings with no prior information about $K$. Using the Pulp-and-paper mill dataset, we investigated the $K$ dependence of retrieval accuracy. Fig. 9 shows MAP@10 versus $K$. The accuracy of DDMM tended to decrease at $K \geq 5$. In contrast, the accuracy of AE + DDMM tended to decrease at $K \geq 15$ and the trend was more gradual than that of DDMM. AE+DDMM was more robust. We think that this is because the AE captured features of segments that become increasingly complex as $K$ increases. This is a desirable characteristic for efficient selection of $K$.

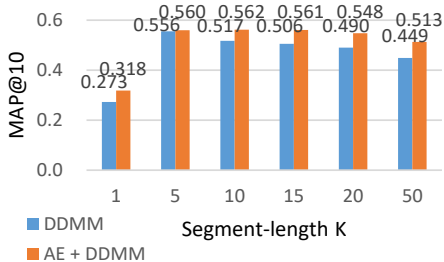

Fig. 9. MAP@10 vs. segment-length $K$ for Pulp-and-paper mill dataset.

## VI. CONCLUSIONS

We propose DDMM to improve retrieval accuracy in unsupervised MTS similarity retrieval. DDMM enables learning of minute differences within states in the entire time series, and thereby recognition of minute differences between states, which are of interest to users in industrial plants. Our empirical studies showed that DDMM significantly outperformed state-of-the-art methods on the Pulp-and-paper mill dataset and demonstrated the effectiveness of DDMM in industrial plants. Furthermore, through AE + DDMM, we showed that accuracy can be further improved by linking DDMM with existing feature extraction methods.

REFERENCES

[1] D. Zhu, D. Songg, Y. Chen, C. Lumezanu, W. Cheng, B. Zong, J. Ni, T. Mizoguchi, T. Yang, and H. Chen, "Deep Unsupervised Binary Coding Networks for Multivariate Time Series Retrieval", *AAAI*, vol. 34, no. 02, pp. 1403-1411, Apr. 2020.

[2] Z. Tan, M. Zhao, Y. Wang, and W. Yang, "Multivariate Time Series Retrieval with Binary Coding from Transformer," 29th International Conference on Neural Information Processing (ICONIP 2022), Nov. 2022. Neural Information Processing, Communications in Computer and Information Science, vol. 1791, pp. 397-408, Apr. 2023.

[3] P. Trirat, Y. Shin, J. Kang, Y. Nam, J. Na, M. Bae, J. Kim, B. Kim, and J-G. Lee, "Universal Time-Series Representation Learning: A Survey," 2024, arXiv:2401.03717. [Online]. Available: https://arxiv.org/abs/2401.03717

[4] C. Wang, X. Li, T. Zhou, and Z. Cai, "Unsupervised Time Series Segmentation: A Survey on Recent Advances," Computers, Materials and Continua, vol. 80, issue 2, pp. 2657-2673, Aug. 2024.

[5] C. Wang, K. Wu, T. Zhou, and Z. Cai, "Time2State: An Unsupervised Framework for Inferring the Latent States in Time Series Data", Proc. ACM Manag. Data 1, 1, Article 17, May 2023.

[6] Z. Lai, H. Li, D. Zhang, Y. Zhao, W. Qian, and C. S. Jensen, "E2USD: Efficient-yet-effective Unsupervised State Detection for Multivariate Time Series," Proc. of the ACM Web Conference 2024 (WWW '24), pp. 3010–3021, May 2024.

[7] J-Y. Franceschi, A. Dieuleveut, and M. Jaggi, "Unsupervised Scalable Representation Learning for Multivariate Time Series," Proc. of the 33rd International Conference on Neural Information Processing Systems, Article 418, pp. 4650–4661, Dec. 2019.

[8] S. Tonekaboni, D. Eytan, and A. Goldenberg, "Unsupervised Representation Learning for Time Series with Temporal Neighborhood Coding," 9th International Conference on Learning Representations (ICLR 2021), May 2021.

[9] G. E. Hinton and R. R. Salakhutdinov, "Reducing the Dimensionality of Data with Neural Networks", Science, vol. 313, Paper no. 5786, pp. 504-507, July 2006.

[10] E. Eldele, M. Ragab, Z. Chen, M. Wu, C. K. Kwoh, X. Li, and C. Guan, "Time-Series Representation Learning via Temporal and Contextual Contrasting," Proc. of the Thirtieth International Joint Conference on Artificial Intelligence (IJCAI-21), pp. 2352-2359, Aug. 2021.

[11] Z. Yue, Y. Wang, J. Duan, T. Yang, C. Huang, Y. Tong, and B. Xu, "TS2Vec: Towards Universal Representation of Time Series", *AAAI*, vol. 36, no. 8, pp. 8980-8987, Jun. 2022.

[12] C. Ranjan, M. Reddy, M. Mustonen, K. Paynabar, and K. Pourak, "Dataset: Rare Event Classification in Multivariate Time Series," 2019, arXiv:1809.10717. [Online]. Available: https://arxiv.org/abs/1809.10717

[13] N. Srivastava, E. Mansimov, and R. Salakhutdinov, "Unsupervised Learning of Video Representations Using LSTMs," Proc. of the 32nd International Conference on International Conference on Machine Learning (ICML'15), vol. 37, pp. 843–852, July 2015.

[14] S. Naito, Y. Taguchi, K. Nakata, and Y. Kato, "Anomaly Detection for Multivariate Time Series on Large-scale Fluid Handling Plant Using Two-stage Autoencoder," 2021 International Conference on Data Mining Workshops (ICDMW), pp. 542-551, Dec. 2021.

APPENDIX. DATA CLEANSING OF EEG EYE STATE DATASET

Fig. A1 shows the trend graphs of all 14 sensors in the EEG eye state dataset. The graph '15 eyeDetection' shows label information ('1' for the eye-closed state and '0' for the eye-open state). The sensor values include large pulse signals, presumably measurement noise, which could interfere with the performance evaluation. Therefore, we excluded those pulse signals by removing rows in the data where any of the sensor values were less than 3000 or greater than 5000. Figure A2 shows the trend graphs after removal. These removed data are used in our experiments.

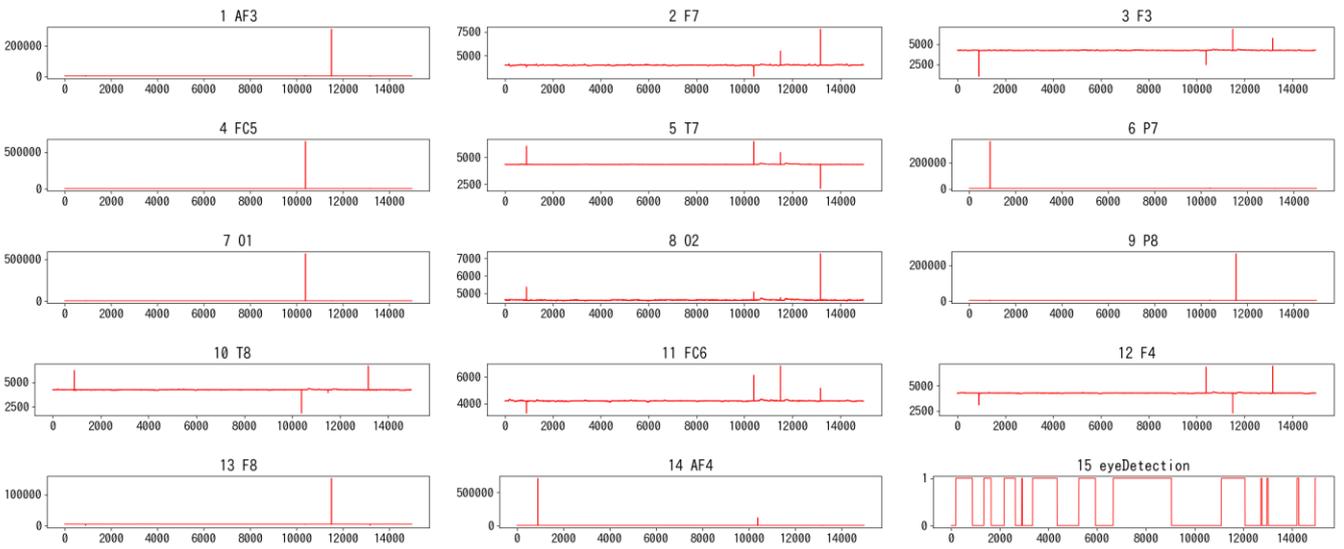

Fig.A1. Trend graphs of all 14 sensors in EEG eye state dataset.

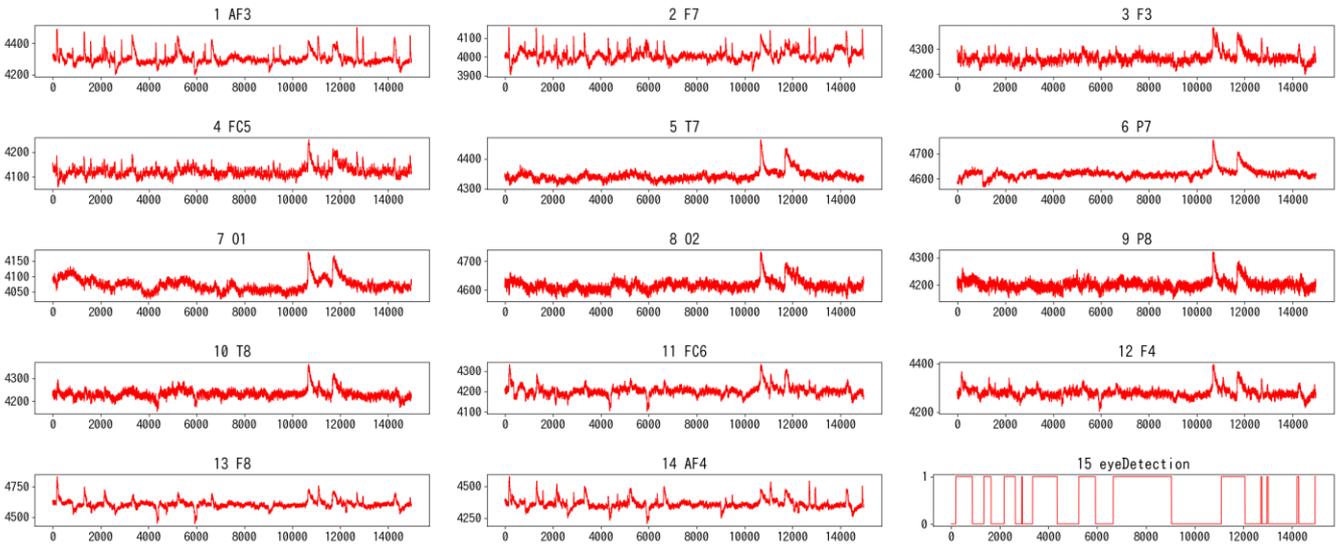

Fig.A2. Trend graphs after removing large pulse signals.